\documentclass[runningheads]{llncs}
\usepackage{graphicx}
\usepackage{amsmath,amssymb} 
\usepackage{color}
\usepackage[width=122mm,left=12mm,paperwidth=146mm,height=193mm,top=12mm,paperheight=217mm]{geometry}

\usepackage{subfigure}

\usepackage{algorithm}
\usepackage{algorithmic}
\usepackage{mathrsfs}

\begin{document}
\pagestyle{headings}
\mainmatter
\def\ECCV16SubNumber{}  

\title{Deep Feature Based Contextual Model for Object Detection} 




\author{Wenqing Chu and Deng Cai}
\institute{State Key Lab of CAD \& CG, Zhejiang University}

\maketitle

\begin{abstract}
	
	Object detection is one of the most active areas in computer vision, which has made significant improvement in recent years. 
	Current state-of-the-art object detection methods mostly adhere to the framework of regions with convolutional neural network (R-CNN) and only use local appearance features inside object bounding boxes. Since these approaches ignore the contextual information around the object proposals, the outcome of these detectors may generate a semantically incoherent interpretation of the input image. 
	In this paper, we propose an ensemble object detection system which incorporates the local appearance, the contextual information in term of relationships among objects and the global scene based contextual feature generated by a convolutional neural network. The system is formulated as a fully connected conditional random field (CRF) defined on object proposals and the contextual constraints among object proposals are modeled as edges naturally. Furthermore, a fast mean field approximation method is utilized to inference in this CRF model efficiently.
	The experimental results demonstrate that our approach achieves a higher mean average precision (mAP) on PASCAL VOC 2007 datasets compared to the baseline algorithm Faster R-CNN.

\keywords{Object Detection, Context Information, Conditional Random Field}
\end{abstract}

\section{Introduction}

Object detection is one of the fundamental problems in computer vision. It plays an important role in many real-world applications such as image retrieval, advanced driver assistance system and video surveillance. This problem is very difficult because the object appearances vary dramatically from changes in different illuminations, view points, nonrigid deformations, poses, and the presence of occlusions. For instance, there is a large amount of partial occlusions between pedestrians standing next to each other in a crowd street and birds come in various poses and colors.

In the past few years, remarkable progress has been made to boost the performance of object detection. A common pipeline to address this problem consists of two main steps: (1) object proposal generation, (2) class-specific scoring and bounding box regression. There is a significant body of methods for generating object proposals such as \cite{carreira2012cpmc,uijlings2013selective,arbelaez2014multiscale,krahenbuhl2014geodesic,zitnick2014edge} or just a sliding window fashion \cite{felzenszwalb2010object}. Then some specific feature of the object bounding box is extracted and some classifier is applied for efficient and accurate object detection, in which the representative work includes AdaBoost algorithm \cite{viola2004robust}, DPM model \cite{felzenszwalb2010object} and deep CNN models \cite{girshick2014rich}. However, most state-of-the-art detectors like Faster RCNN \cite{ren2015faster} only consider the proposals individually without taking the contextual information into account.

In the real world, there exists a semantic coherent relationship between the objects both in terms of relative spatial location and co-occurrence
 probability \cite{biederman1972perceiving,biederman1982scene}. In some situations, contextual information among objects in the input image can provide a more valuable cue for the detection of an object than the information near an object’s region of interest. In addition, the global context based on scene understanding also helps the detector better rule out the false alarm.
    
        \begin{figure}[h]
        	\begin{center}
        		\subfigure[Boat and Train]{
        			\label{fig:subfig:aa} 
        			\includegraphics[scale=0.35]{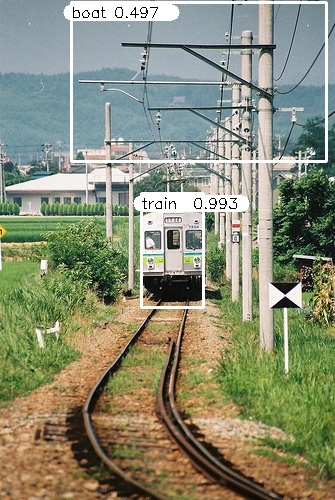}}
        		\subfigure[Partial Occlusions between people]{
        			\label{fig:subfig:ab} 
        			\includegraphics[scale=0.35]{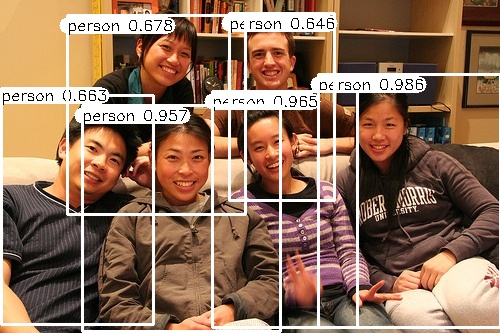}}
        		\caption{Object-level contextual information}
        		\label{fig:subfiga} 
        	\end{center}
        \end{figure}

              \begin{figure}[h]
              	\begin{center}
              		\subfigure[Boats in Lake]{
              			\label{fig:subfig:ba} 
              			\includegraphics[scale=0.33]{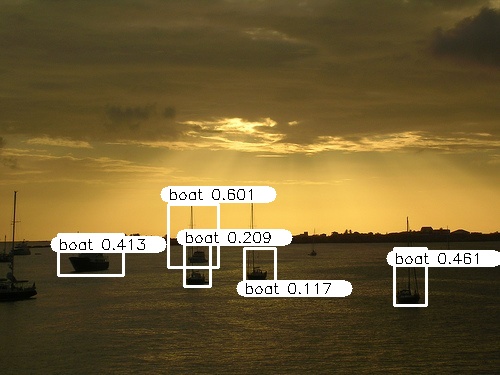}}
              		\subfigure[Aeroplanes in air]{
              			\label{fig:subfig:bb} 
              			\includegraphics[scale=0.33]{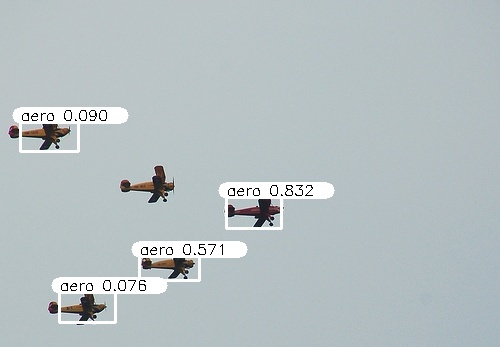}}
              		\caption{Image-level contextual information}
              		\label{fig:subfigb} 
              	\end{center}
              \end{figure}

    In Figure~\ref{fig:subfig:aa}, Faster R-CNN recognizes the two object proposals individually. However, boats and trains stand little chance of co-occurence in the input image , which means the probability for the boat or the train should be decreased. In Figure~\ref{fig:subfig:ab}, since the person is occluded by the sofa and Faster R-CNN classify the object proposal into person by the probability of $0.646$. However, if we can use the contextual information of objects around the bounding box to help inference, we can raise our confidence that the category of this object candidate behand the sofa is a person. In addition, image-level contextual information can also support object detection. If we can recognize the scene in Figure~\ref{fig:subfig:ba} as a lake or sea, then we can make a better judgement that the object proposals are likely to be tiny boats. Similarily, apperance in Figure~\ref{fig:subfig:bb} looks like sky which enhances the probability of the presence of aeroplanes.
    In a word, the context information can be a strong clue for recognizing the object candidates which are ambiguous because of low resolution and variation in pose and illumination.

    In the history of computer vision, a number of approaches \cite{torralba2003contextual,crandall2007composite,rabinovich2007objects,tu2008auto,galleguillos2008object,song2011contextualizing,choi2010exploiting,desai2011discriminative,mottaghi2014role} have exploited contextual information in order to boost the preformance of object detection.
     Nevertheless, most of these methods leverage hand-crafted features such as Gist \cite{torralba2003contextual} or HOG \cite{dalal2005histograms} to extract features from the input image. Recently, convolutional neural network (CNN) has achieved great success in computer vision tasks such as image classification \cite{krizhevsky2012imagenet}, which inspired us to employ CNN to devise a novel contextual model.

 In this paper, we propose a novel context model which is based on the prominent object detection method Faster R-CNN \cite{ren2015faster} in recent years.    
  To leverage the context information around each candidate, we first focus on the local contextual classes that are present near the object proposal. As in \cite{galleguillos2008object}, inter-object constraints vary greatly from changes in different categories and locations, which can be learned from the statistical summary of the datasets. In addition, we also generate the global scene descriptor using a CNN model which is trained for the secne understanding task. Then the global scene information of the input image is utilized as the input of a logistic regression method to predict the probability how much some category is likely to occur in the image. In the following step, we apply Faster R-CNN to each input image and obtain a pool of object proposals with the corresponding scores and locations for each category. After that, we take these object proposals as nodes in graph and formulate the graph as a fully-connected conditional random field (CRF) according to the context information. Specifically, the unary potentials are determined by the scores from Faster R-CNN. And the pairwise postentials are decided by the layouts and categories among the object proposals. To efficiently inference in this CRF, we utilize a fast mean field inference algorithm of \cite{krahenbuhl2012efficient,hayder2014object} to yield the candidate labels and the corresponding confidence simultaneously.

We have extensively evaluated our method on the PASCAL VOC2007 dataset \cite{everingham2008pascal}. The experimental results show that our method can acheieve an improvement of $0.87\%$ in mAP and for the category bottle the AP can be raised up to $3.4\%$. Very few categories’ accuracies are worsened by context information.

The rest of this paper is arranged as follows. First, we briefly review a few of recent work on object detection methods in Section~\ref{section:related}. Then in section~\ref{section:crf} we describe the framework of our context model for object detection and the inference algorithm in detail. After that, we evaluate the performance of our method on the challenging databases PASCAL 2007 in section~\ref{section:experiments}. Finally, in section~\ref{section:conclusions}, we present our conclusions and discuss future work.

\section{Related Work}\label{section:related}

In this section, we briefly review the recent work on object detection. Object detection has been active research areas in recent years, which has lead to a large amount of methods to address the problems in it.

In the literature of object detection, the part-based model is one of the most powerful approaches in which deformable part-based model (DPM) \cite{felzenszwalb2010object} is an excellent example. This method utilize mixtures of multiscale deformable part models to represent highly variable objects.
 Each part captures local appearance properties of an object while the deformable configuration is characterized by spring-like connections between certain pairs of parts. It detect objects in an image by a sliding window approach and take the histogram of oriented gradients (HOG) features \cite{dalal2005histograms} as input.

Recently, deep convolutional neural networks (CNN) have emerged as a powerful machine learning model on a number of image recognition benchmarks, including the most noticeably work by \cite{krizhevsky2012imagenet}. That aroused a significant body of methods
 \cite{szegedy2013deep,sermanet2013overfeat,erhan2014scalable,girshick2014rich,he2015spatial,girshick2015fast,ren2015faster,bell2015inside,liu2015box,gidaris2015locnet,redmon2015you,he2015deep} addressing the problem with CNN. Among these approaches, the regions-with-convolutional-neural-network (R-CNN) framework \cite{girshick2014rich} achieved excellent detection performance and becomes a commonly employed paradigm for object detection. Its essential steps include proposal generation with selective search \cite{uijlings2013selective}, CNN feature extraction, object box classification and regression based on the CNN feature. However, R-CNN brings excessive computation cost because it extracts CNN feature repeatedly for thousands of object proposals. Spatial pyramid pooling networks (SPPnets) \cite{he2015spatial} were proposed to accelerate the process of feature extraction in R-CNN by sharing the forward pass computation.The SPPnet approach computes a convolutional feature map for the entire input image once and then generates a fixed-length feature vector from the shared feature map for each object proposal. Fast Region-based Convolutional Network method (Fast R-CNN) \cite{girshick2015fast} utilizes a multi-task loss,which leads to an end-to-end framework that the training is a single-stage and no disk storage is required for feature caching. The drawback of Fast R-CNN is that this method still use bottom-up proposal generation which is the bottleneck of efficiency. In \cite{ren2015faster}, the authors proposed a Region Proposal Network (RPN) method that shares full-image convolutional features with the detection network, thus enabling nearly cost-free region proposals.These techniques, however, still mostly perform detection based on local appearance features in the input image.

 In the other hand, semantic context also plays a very important role of boosting the performance of object detection \cite{torralba2003contextual,rabinovich2007objects,galleguillos2008object,choi2010exploiting,desai2011discriminative,mottaghi2014role}. 
 The statistics of low-level features across the entire scene were used to predict the presence
 or absence of objects and their locations \cite{torralba2003contextual}.
 In \cite{rabinovich2007objects}, the authors demonstrated that contextual relations between objects' labels can help reduce ambiguity in objects' visual appearance. Specifically, they utilized image segmentation as a pre-processing step to generate object proposals. Then a conditional random field (CRF) formulation was exploited as post-processing to infer the optimal label configuration of this CRF model, which jointly labels all the object candidates. \cite{galleguillos2008object} extend this approach to combine two types of context – co-occurrence and relative location – with local appearance based features. 
 \cite{desai2011discriminative} introduced a unified model for multi-class object recognition that learns statistics that capture the spatial arrangements of various object classes in real images.

Besides, some work \cite{yao2010modeling,li2014integrating,vu2015context} which focus on detecting some specialized object class were proposed to use context information to support detection. However, these models mostly work on context information represented by traditional visual features such as HOG or GIST. Thus, we are motivated to move on to more powerful features provided by CNN model.

\section{A Fully-Connected CRF for Object Detection}\label{section:crf}

In this section, we address the general object detection problem with an ensemble system, in which we combine the local appearence, the contextual relationships among object candidates and the global scene context information. To process each input image, our approach includes three main stages. At first, we generate a pool of object proposals obtained with Faster R-CNN \cite{ren2015faster}. Then we use a conditional random field (CRF) framework to model the object detection problem with contextual information. Finally, we employ an efficient mean field approximation method to inference and finaly maximize object label agreement.

In Section ~\ref{section:proposal}, we introduce the process of object proposal generation building on Faster R-CNN \cite{ren2015faster}.
In Section ~\ref{section:crff}, We give the formulation of the CRF model for the object detection problem. We will describe unary potentials, pairwise potentials between object candidates and the global potentials determined by the context feature describing the entire image. After all, Section ~\ref{section:learning} will give the inference algorithm.

\subsection{Object Proposals} \label{section:proposal}

Our approach generates object proposals following Faster R-CNN \cite{ren2015faster}, which is one of the state-of-the-art object detection methods. In contrast to R-CNN \cite{girshick2014rich}, Faster R-CNN proposed a Region Proposal Network (RPN) instead of other bottom-up approaches to outputs a set of object bounding boxes. Since RPN slides a small network over the conv feature map output by the last conv layer, which makes it can share forward pass computation with a Fast R-CNN object detection network \cite{girshick2015fast} and that leads to great advantages on efficiency.

Faster R-CNN method is a object detection system which can depend on different CNN architecture. In our experiments, we investigate the Zeiler and Fergus model \cite{zeiler2014visualizing} (ZF), which has 5 shareable conv layers and the Simonyan and Zisserman model \cite{simonyan2014very} (VGG), which has 13 shareable conv layers. To verify our method is insensitive to different object proposal methods, we conduct experiments on both CNN models. To train the network, we optimize parameters with the popular stochastic gradient descent (SGD) \cite{lecun1989backpropagation} with momentum. The Faster R-CNN model is pre-trained on the ImageNet dataset \cite{deng2009imagenet} and fintuned on the PASCAL VOC 2007 dataset \cite{everingham2008pascal}. More details on the training procedure can be found in \cite{ren2015faster}.

\subsection{CRF Formulation} \label{section:crff}

In this stage, we take each object proposal generated by the first step in the pipeline as a node.
Suppose there are $N$ proposals in a single input image and $K$ categories, we can consider a random field $\mathbf{X}$ defined over a set of variables $\{x_1,\dots,x_N\}$. The domain of each variable is a set of categories $L = \{0, 1, 2, \dots, K \}$ in which $0$ represents background and $x_j$ is the label assigned to proposal $j$. Faster R-CNN generates a scores matrix $\mathbf{S} \in \mathbb{R}^{N\times K}$ which $\mathbf{S}_{j,k}$ represents the probability of proposal $j$ belong to category. These scores are used as initial scores of each node. Our method will adjust the scores of each node in a fully connected graph according to the scene information and the contextual relationship among the nodes. 
  
 \subsubsection{Unary Potentials} \label{section:local}
 In our conditional random field (CRF) model, the unary potential $\phi_u(x_i)$ measures the probability that the object proposal $i$ belongs to the category $x_i$ according to the local appearance. Therefore, we use a rescaled score from Faster R-CNN as unary potential. This lets us write our unary term as
 \begin{equation}\label{eq:unary}
 \phi_u(x_i) = - log (\mathbf{S}_{i,x_i})
 \end{equation}
 where $\mathbf{S}_{i,x_i}$ is the confidence that proposal $i$ belong to category $x_i$.

\subsubsection{Pairwise Potentials} \label{section:pairwise}

\begin{figure}[h]
              	\begin{center}
              		\subfigure[Pottedplants]{
              			\label{fig:subfig:ca} 
              			\includegraphics[scale=0.3]{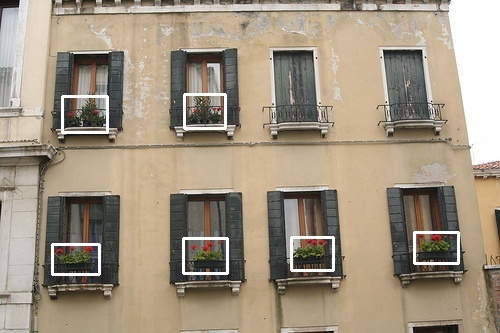}}
              		\subfigure[Horse and Human]{
              			\label{fig:subfig:cb} 
              			\includegraphics[scale=0.3]{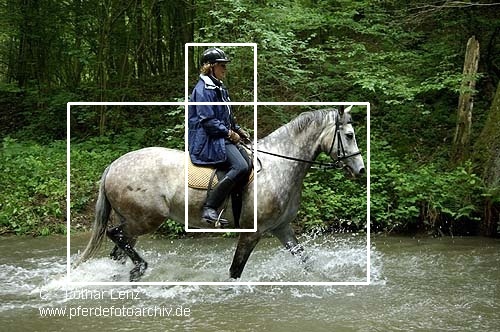}}
              		\caption{different spatial relationship}
              		\label{fig:subfigc} 
              	\end{center}
              \end{figure}

Here we describe our pairwise potentials that purpose to capture the contextual information between multiple object candidates. Our pairwise model takes both the semeantic and spatial relatioinships into account like \cite{galleguillos2008object,desai2011discriminative}.
In our method, we define a function $\phi_p()$ to estimate the pairwise potential of a proposal $i$ based on a known object $j$ in the same image. The input of $\phi_p()$ is the category information of $i$ and $j$ and their relative location. To leverage this spatial relevance better, we consider $11$ different layouts for two object bounding boxes. If two candidates dont't have intersection, then the spatial relationships of them can be classified into far, up, down, left and right. Otherwise, they can be defined by inside, outside, up, down, left and right. For example, some pottedplants are above another in Figure~\ref{fig:subfig:ca} and don't intersect between each other. InFigure~\ref{fig:subfig:cb}, however, the human is on the horse and their bounding boxes have intersection. We define these situations as two different spatial relationships.

 Following \cite{galleguillos2008object,desai2011discriminative}, this function can be defined according to the likelihood $P(x_i,x_j, r)$ which is learned from statistic summary of the training dataset.

\begin{equation}\label{eq:pairwise}
\phi_p(x_i, x_j, r) = -log ( P(x_i,x_j, r))
\end{equation}
where $P(x_i,x_j, r)$, with r = 1,...,11, measures the likelyhood that an object with label $x_i$ appears with an object label $x_j$ for a given relationship $r$.

\subsubsection{Global Potentials} \label{section:global}

In addition to object-level contextual information, we also introduce image-level signal to reason about presence or absence of objects in the image. The key point is to find global image features which can represent various secne well.

Scene categorization is also a challenge task in computer vision. Before, most work focus on shallower hand-crafted features empirically and the databases that they used are lack of abundance and variety. Recently, the Places2 dataset \cite{zhou2015places2} is provided which contains more than 10 million images comprising 400+ unique scene categories. Moveover, the dataset features 5000 to 30,000 training images per class, consistent with real-world frequencies of occurrence.
With this large dataset, we can apply the powerful CNN model to feature extraction representing the scene context information. Specifically,  the CNN model takes the whole image as input and outputs a score for each category. Following \cite{zhou2014learning}, we train the VGG network \cite{simonyan2014very} on the dataset using Caffe deep learning toolbox. And the last layer is used to represent the scene context. With the generic deep scene features for visual recognition, we use a logistic regression model to fit it and output $p(x_i|f)$ which measures the probability of existence of the category $x_i$ in the input image.

\begin{equation}\label{eq:global}
\phi_g(x_i) = -log ( p(x_i|f))
\end{equation}
where $(x_i|f)$ measures the likelyhood that an object with label $x_i$ appears in the input image.

Then we formulate the fully-connected CRF model as below: 

\begin{equation}\label{eq:crf}
E(\mathbf{X}) = \sum_{i=1}^{N} \phi_u(x_i) + \omega_{p}\sum_{i<j} \phi_p(x_i,x_j, r) + \omega_{g}\sum_{i=1}^{N} \phi_g(x_i)
\end{equation}
 where $\omega_{p}$ and $\omega_{g}$are the weights of the pairwise potentials and the global potentials.

\subsection{Inference Algorithm} \label{section:learning}

To tackle this fully-connected CRF model, we use the mean field approximation method to minimize the objective function. Following \cite{krahenbuhl2012efficient}, we adopt a fast mean field approximation algorithm to compute the marginals. Given the current mean field estimates $\{Q_i\}$ of the marginals, the update equation can be written as

\begin{equation}\label{eq:mf}
Q_i(x_i) \propto exp( - \phi_u(x_i) - \omega_{g}\phi_g(x_i) - \omega_{p}\sum_{x_j \neq x_i} \sum_{j \neq i} Q_j(x_j) \phi_p(x_j,x_i))
\end{equation}

After convergence, we obtain an (approximate) posterior distribution of object labels for each node. To obtain the final results, we can employ the mean field approximate marginal probability $Q_i(x_i)$ as a detection score. Since the number of object proposals is mostly around $300$, the time cost is almost free. The whole procedure for inference has been presented in Algorithm \ref{algo:meanfield}..

\begin{algorithm}
	
	\caption{Mean field in fully connected CRFs}
	\begin{algorithmic}[1]
		\STATE Initialize $Q$
		\[
		\ Q_i(x_i) = \frac{1}{Z_i} exp\{-\phi_u(x_i) - \omega_{g}\phi_g(x_i) \}
		\]		   		
		
		\WHILE {not converged} 
		
		\STATE Message passing.
		\[
		\ \tilde{Q_i(l)} = \sum_{j \neq i} Q_j(l)
		\]	
		
		\STATE Compatibility transform
		\[
		\ \hat{Q_i(x_i)} = \sum_{l \in L} \tilde{Q_i(l)} \omega_{p}\phi_p(l,x_i,r)
		\]	
		
		\STATE Local update 
		\[
		\ Q_i(x_i) = \frac{1}{Z_i} exp\{-\phi_u(x_i) - \omega_{g}\phi_g(x_i) - \hat{Q_i(x_i)} \}
		\]
		
		\ENDWHILE
	\end{algorithmic}\label{algo:meanfield}
\end{algorithm}

\section{Experiments}\label{section:experiments}

In this section, we conduct a series of experiments to evaluate the performance of our approach and compare it against the state-of-the-art object detection baseline. We evaluate our method on object detection benchmark datasets PASCAL VOC 2007 \cite{everingham2008pascal}. There are 20 different categories of objects and every dataset is divided into train, val and test subsets. In this dataset, object appearances vary greatly from changes in different illuminations, poses, locations, viewpoints and the presence of occlusions. We compare the performance in terms of mean average precision (mAP) which is the principal quantitative measure in VOC object detection task \cite{everingham2008pascal}. The results demonstrate that our method can boost the detection performance based on the baseline method.

\subsection{Experiment and Evaluation Details}

All of our experiments use the Faster R-CNN method \cite{ren2015faster} as our baseline. In order to show that our method is insensitive to the stage of object proposal, we utilize two CNN architecture including ZFNet \cite{zeiler2014visualizing} and VGGNet \cite{simonyan2014very}  to train the Faster R-CNN system. In addition, we use different traing data to obtain different CNN model. In what follows, we use ZF to denote Faster R-CNN with ZFNet, VGG to denote Faster R-CNN with VGGNet. Pp means the pairwise potentials are added, while Gp means the incorporation of the global potentials. The weight parameters for the pairwise potentials and the global potentials are selected via cross validation. We show the performance in terms of AP for each class on VOC 2007 test.

\subsection{Analysis}
Firstly, we examine the influence of parameters in our model. Then we show the performance in terms of average precision on the VOC 2007 test.

\subsubsection{Parameter Selection}
\begin{figure}[h]
              	\begin{center}
              		\subfigure[AP for Bottle]{
              			\label{fig:subfig:psa} 
              			\includegraphics[scale=0.4]{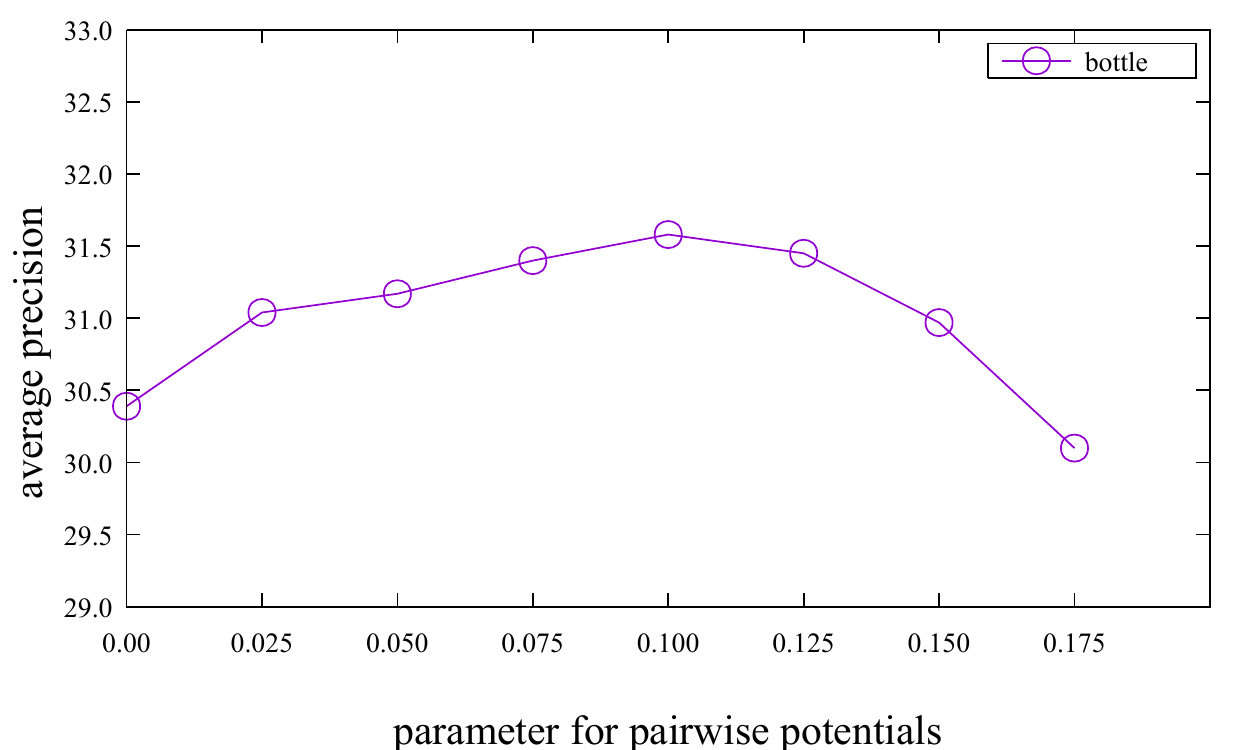}}
              		\hspace{1in}
              		\subfigure[AP for Cow]{
              			\label{fig:subfig:psb} 
              			\includegraphics[scale=0.4]{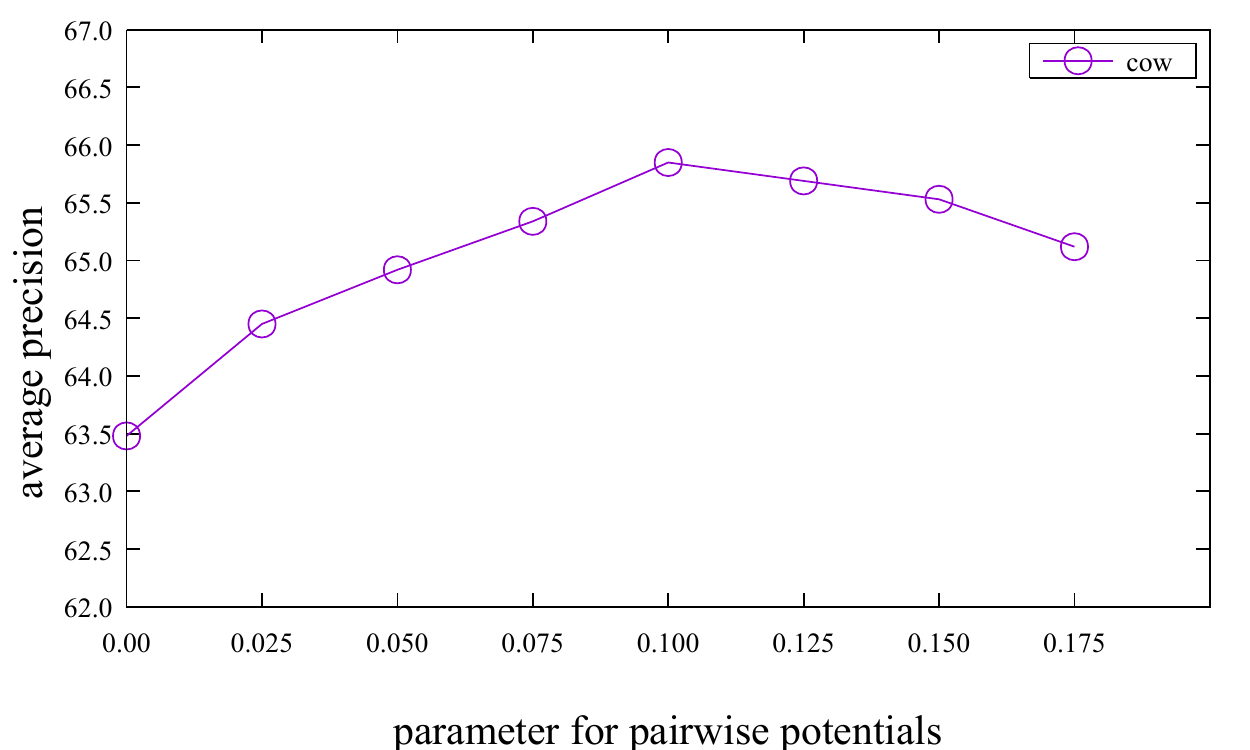}}
              		\subfigure[AP for Pottedplant]{
              			\label{fig:subfig:psc} 
              			\includegraphics[scale=0.4]{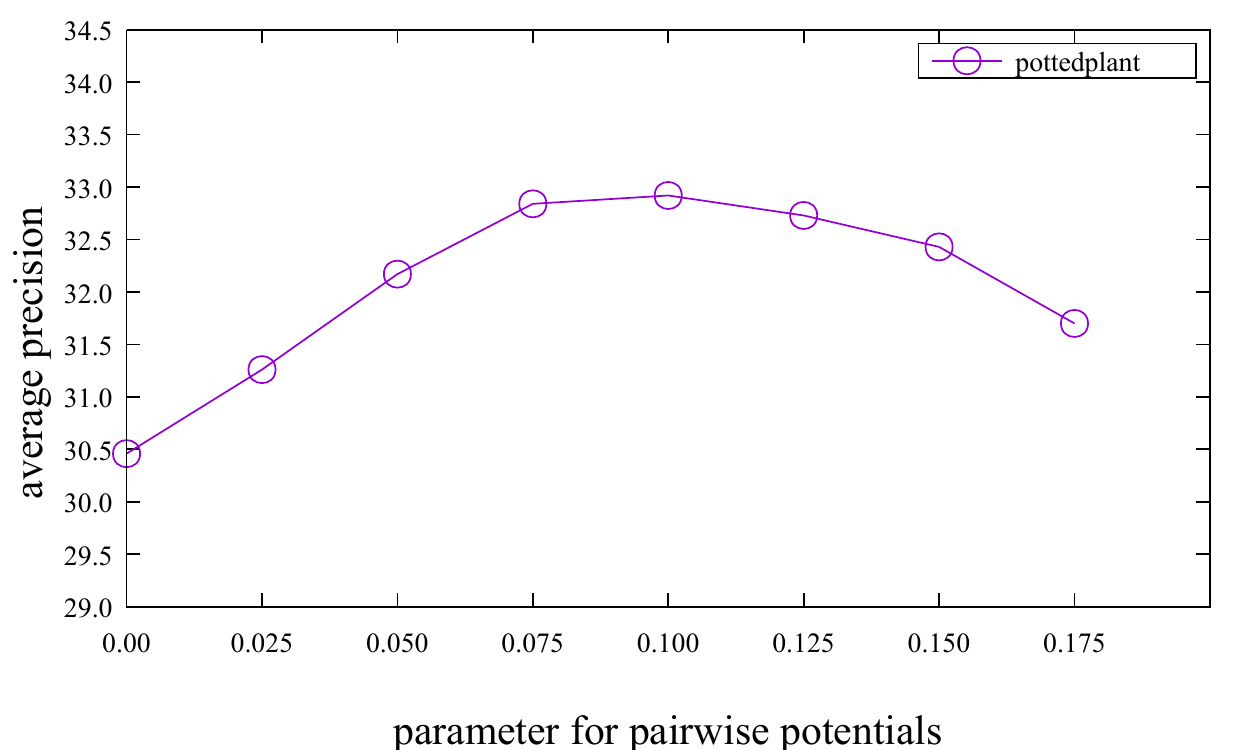}}
              		\caption{AP for three classes with different $\omega_{p}$}
              		\label{fig:subfigps} 
              	\end{center}
              \end{figure}

In our CRF framework, the parameter $\omega_{p}$ and $\omega_{g}$ in \eqref{eq:crf} adjust the tradeoff between the local apperance and the contextual information. Here we focus on $\omega_{p}$ which represents the importance of object-level contextual information and other hyperparameters are fixed. 
The results of average precision for classes bottle, cow and pottedplants are shown in Figure~\ref{fig:subfigps}. As shown in Figure\ref{fig:subfigps}, when the parameter $\omega_{p}$ becoms larger, our method achieves better AP at first and then decrease. It demonstrates that there exists a balance for local appearance and contextual coherent constraints among object candidates in the same image.

\subsubsection{performance on the ZFnet}
In Table~\ref{table:zf}, we take ZFnet which is trained on the VOC 2007 trainval dataset as the baseline and our approach obtains $0.87\%$ improvement in terms of mAP. In this experiment, our method yields the best performance over $18$ classes. Among these categories, the class bottle is enhanced by $3.4\%$ and class tvmonitor achieves $2.2\%$ improvement. Both pairwise potentials and global potentials support the object detection. Morerover, we can see that the pairwise potentials part plays a more important role because it achieves improvement in 8 calsses compared to 3 classes' enhancement caused by the global potentials. We show the results of ZFnet trained on the union of VOC 2007 and 2012 trainval datasets in Table~\ref{table:zf0712}.

\setlength{\tabcolsep}{4pt}
\begin{table}
		\caption{The result of VOC 2007 test dataset based on ZFnet trained with VOC2007 trainval dataset}
		\label{table:zf}
		\centering
		\begin{tabular}{|c|c|c|c|c|c|c|c|c|c|c|}  \hline
			class & aero & bike & bird & boat & bottle & bus & car & cat & chair & cow \\ \hline
			ZF & 64.0 & \textbf{69.9} & 56.6 & 44.9 & 30.3 & 66.8 & 73.4 & 71.0 & 35.6 & 63.4 \\ 
			ZF+Pp & 65.4 & 69.6 & \textbf{57.8} & 45.6 & 31.5 & 67.1 & 73.6 & \textbf{71.6} & 36.2 & \textbf{65.8} \\ 
			ZF+Gp & 64.9 & 69.3 & 57.0 & 45.4 & 33.4 & \textbf{67.8} & 73.7 & 70.8 & 36.5 & 64.5 \\ 
			ZF+Pp+Gp & \textbf{65.9} & 69.8 & 57.2 & \textbf{45.6} & \textbf{33.7} & 67.2 & \textbf{73.8} & 71.1 & \textbf{37.3} & 65.0 \\ \hline
			class & table & dog & horse & m-bike & person & plant & sheep & sofa & train & tv \\ \hline
			ZF & 60.2 & 65.5 & 76.7 & \textbf{70.9} & 64.4 & 30.4 & 58.0 & 53.5 & 72.3 & 56.6 \\ 
			ZF+Pp & 58.3 & \textbf{66.5} & 76.6 & 70.6 & \textbf{64.6} & \textbf{32.9} & \textbf{61.0} & 52.6 & \textbf{72.8} & 57.3 \\ 
			ZF+Gp & \textbf{61.5} & 65.1 & 76.7 & 70.8 & 64.0 & 31.0 & 56.6 & \textbf{54.6} & 72.6 & 58.7 \\ 
			ZF+Pp+Gp & 61.4 & 65.6 & \textbf{77.2} & 70.3 & 64.1 & 31.8 & 59.4 & 54.2 & 72.6 & \textbf{58.8} \\ \hline
			
		\end{tabular}
\end{table}
\setlength{\tabcolsep}{1.4pt}

\setlength{\tabcolsep}{4pt}
\begin{table}
		\caption{The result of VOC 2007 test dataset based on ZFnet trained with VOC2007 and VOC2012 trainval dataset}
		\label{table:zf0712}
		\centering
		\begin{tabular}{|c|c|c|c|c|c|c|c|c|c|c|}  \hline
			class & aero & bike & bird & boat & bottle & bus & car & cat & chair & cow \\ \hline
			ZF & 67.8 & 71.2 & 59.1 & 49.8 & 33.8 & 71.7 & \textbf{75.2} & \textbf{79.9} & 38.4 & 70.4 \\ 
			ZF+Pp & 68.1 & 70.8 & \textbf{60.3} & 49.5 & 35.6 & \textbf{72.1} & 74.9 & 79.3 & 39.3 & \textbf{72.3} \\ 
			ZF+Gp & 68.8 & 70.9 & 60.0 & \textbf{50.2} & 37.6 & 71.7 & 75.1 & 78.9 & 39.1 & 70.4 \\ 
			ZF+Pp+Gp & \textbf{69.2} & \textbf{71.2} & 60.2 & 48.9 & \textbf{38.3} & 71.7 & 75.0 & 79.5 & \textbf{39.7} & 71.8 \\ \hline
			class & table & dog & horse & m-bike & person & plant & sheep & sofa & train & tv \\ \hline
			ZF & 58.9 & 74.1 & 79.8 & 72.5 & 65.0 & 30.3 & 66.4 & 60.6 & \textbf{72.4} & 58.5 \\ 
			ZF+Pp & 59.8 & 74.0 & \textbf{80.4} & 71.7 & \textbf{65.5} & \textbf{33.1} & \textbf{67.6} & 59.1 & 72.3 & 59.2 \\ 
			ZF+Gp & 60.3 & 73.5 & 79.8 & 72.6 & 65.2 & 32.1 & 66.5 & \textbf{61.1} & 72.1 & 59.6 \\ 
			ZF+Pp+Gp & \textbf{60.3} & \textbf{74.2} & 79.5 & \textbf{73.2} & 65.3 & 32.3 & 67.5 & 61.0 & 72.3 & \textbf{60.0} \\ \hline
			
		\end{tabular}
\end{table}
\setlength{\tabcolsep}{1.4pt}

\subsubsection{performance on the VGGnet}

From Table~\ref{table:vgg}, we can see that the result of our method based on VGGnet is $70.18\%$ in terms of mAP which acchives $0.52\%$ improvement. And it outperforms other alternatives on 18 classes out of a total of 20 of them. Furthermore, the result demonstrates that our method have the potential to work similarly well on different object detection methods since it only needs the proposals and the corresponding scores. However, the improvement becomes lower than it achieved by the model based on ZFnet and the reason maybe that the VGGnet CNN model which focuses on candidates themselves are powerful enough to overlook the help from contextual information in some extent. We show the results of VGGnet trained on the union of VOC 2007 and 2012 trainval datasets in Table~\ref{table:vgg0712}.

\setlength{\tabcolsep}{4pt}
\begin{table}
	\caption{The result of VOC 2007 test dataset based on VGGnet}
	\label{table:vgg}
	\centering
	\begin{tabular}{|c|c|c|c|c|c|c|c|c|c|c|}  \hline
		class & aero & bike & bird & boat & bottle & bus & car & cat & chair & cow \\ \hline
		VGG & 69.1 & 78.8 & 67.7 & \textbf{54.8} & 49.4 & 78.1 & 79.9 & \textbf{84.6} & 50.6 & 74.2 \\ 
		VGG+Pp & \textbf{69.9} & 78.8 & 68.6 & 54.3 & 49.3 & \textbf{78.8} & 79.9 & 84.3 & 51.2 & \textbf{77.2} \\ 
		VGG+Gp & 69.1 & 78.7 & 67.8 & 53.7 & \textbf{51.6} & 78.2 & \textbf{80.0} & 84.4 & \textbf{51.3} & 74.8 \\ 
		VGG+Pp+Gp & 69.1 & \textbf{78.8} & \textbf{69.6} & 53.2 & 51.0 & 78.4 & 79.9 & 84.5 & 51.0 & 76.8 \\ \hline
		class & table & dog & horse & m-bike & person & plant & sheep & sofa & train & tv \\ \hline
		VGG & 65.5 & 81.1 & 83.6 & 77.0 & 75.7 & 38.4 & 70.1 & 66.9 & 80.6 & 66.0 \\ 
		VGG+Pp & 64.9 & 81.2 & \textbf{84.5} & \textbf{77.6} & \textbf{75.8} & 40.5 & \textbf{71.2} & 65.8 & 80.6 & 66.3 \\ 
		VGG+Gp & \textbf{67.0} & 81.2 & 84.1 & 76.7 & 75.6 & 40.2 & 70.0 & \textbf{67.4} & 80.7 & \textbf{67.7} \\ 
		VGG+Pp+Gp & 66.7 & \textbf{81.2} & 84.3 & 77.5 & 75.6 & \textbf{40.6} & 70.9 & 65.7 & \textbf{80.9} & 67.0 \\ \hline
		
	\end{tabular}
\end{table}
\setlength{\tabcolsep}{1.4pt}

\setlength{\tabcolsep}{4pt}
\begin{table}
	\caption{The result of VOC 2007 test dataset based on VGGnet trained with VOC2007 and VOC2012 trainval dataset}
	\label{table:vgg0712}
	\centering
	\begin{tabular}{|c|c|c|c|c|c|c|c|c|c|c|}  \hline
		class & aero & bike & bird & boat & bottle & bus & car & cat & chair & cow \\ \hline
		VGG & 75.4 & 79.8 & 74.5 & 59.9 & 52.7 & 82.9 & 84.6 & \textbf{88.3} & 52.5 & 79.2 \\ 
		VGG+Pp & 75.9 & 80.0 & \textbf{75.3} & \textbf{59.9} & 53.9 & \textbf{83.0} & 84.7 & 87.9 & 53.3 & \textbf{83.4} \\ 
		VGG+Gp & 76.3 & 80.0 & 74.6 & 58.0 & 55.6 & 82.3 & 84.6 & 87.8 & 51.3 & 74.8 \\ 
		VGG+Pp+Gp & \textbf{76.3} & \textbf{80.3} & 75.2 & 57.8 & \textbf{56.2} & 82.1 & \textbf{84.9} & 87.6 & \textbf{53.9} & 82.2 \\ \hline
		class & table & dog & horse & m-bike & person & plant & sheep & sofa & train & tv \\ \hline
		VGG & 66.0 & 84.8 & 85.0 & \textbf{76.8} & 76.6 & 36.8 & 75.6 & \textbf{73.1} & 81.7 & 71.5 \\ 
		VGG+Pp & 65.7 & \textbf{85.4} & \textbf{85.5} & 76.6 & \textbf{77.0} & 40.2 & 75.9 & 71.5 & \textbf{82.2} & \textbf{71.5} \\ 
		VGG+Gp & 66.9 & 84.9 & 84.0 & 76.5 & 76.5 & 40.3 & 75.6 & 72.7 & 82.1 & 71.1 \\ 
		VGG+Pp+Gp & \textbf{67.1} & 85.3 & 85.2 & 76.7 & 76.6 & \textbf{40.4} & \textbf{76.0} & 72.4 & 82.0 & 71.0 \\ \hline
		
	\end{tabular}
\end{table}
\setlength{\tabcolsep}{1.4pt}

\subsection{Visualization of More Results}

\begin{figure}[h]
        	\begin{center}
        			\includegraphics[scale=0.22]{images/2_z.jpg}
        			\includegraphics[scale=0.22]{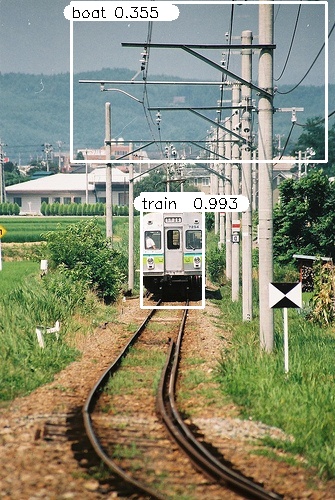}
        			\includegraphics[scale=0.22]{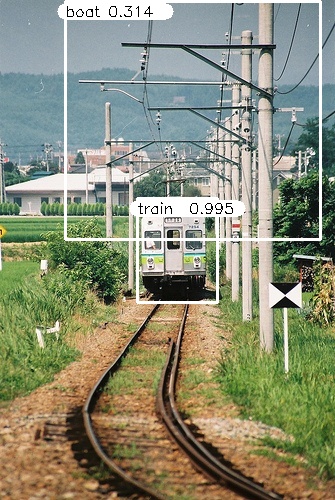}
        			\includegraphics[scale=0.22]{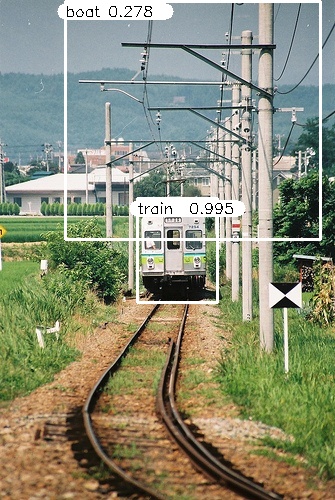}

        			\includegraphics[scale=0.15]{images/4_z.jpg}
        			\includegraphics[scale=0.15]{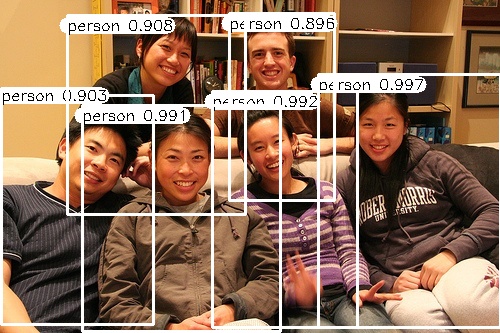}
        			\includegraphics[scale=0.15]{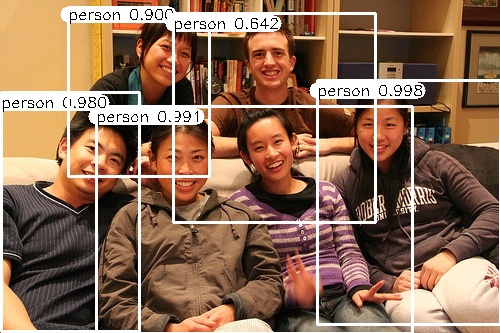}
        			\includegraphics[scale=0.15]{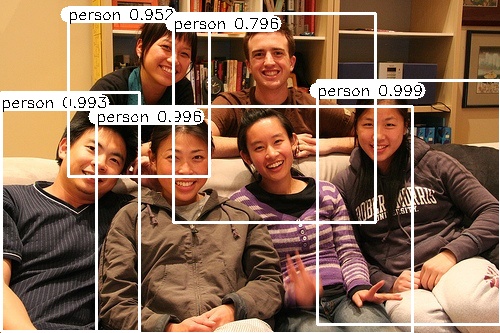}
        		
        			\includegraphics[scale=0.15]{images/6_z.jpg}
        			\includegraphics[scale=0.15]{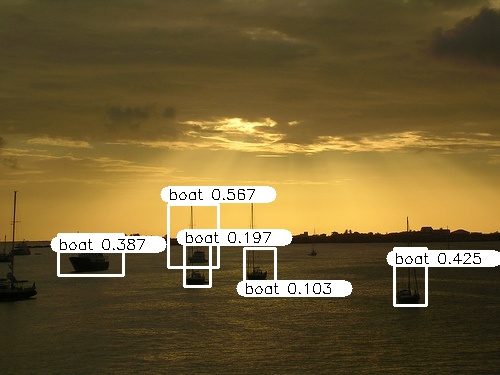}
        			\includegraphics[scale=0.15]{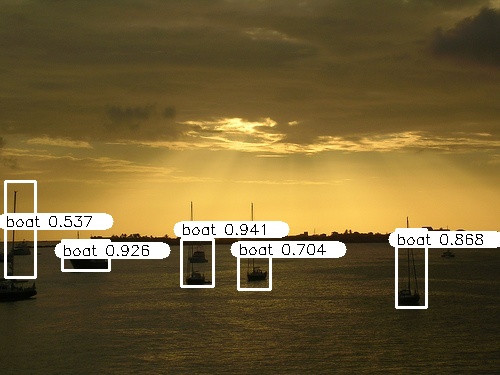}
        			\includegraphics[scale=0.15]{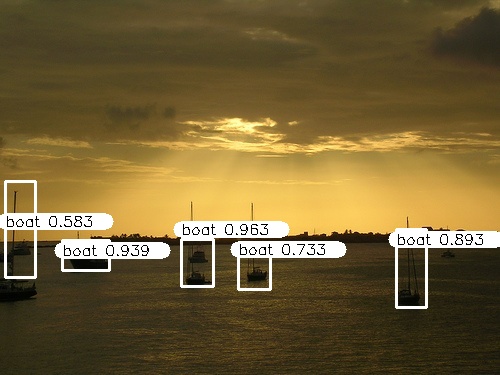}
        		
        			\includegraphics[scale=0.15]{images/8_z.jpg}
        			\includegraphics[scale=0.15]{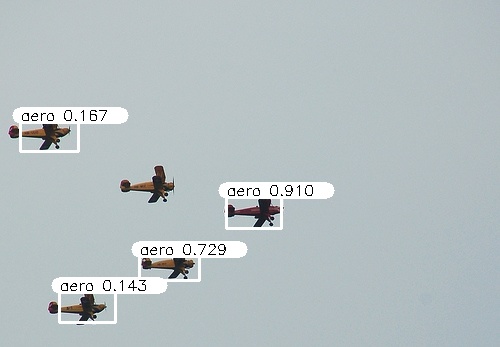}
        			\includegraphics[scale=0.15]{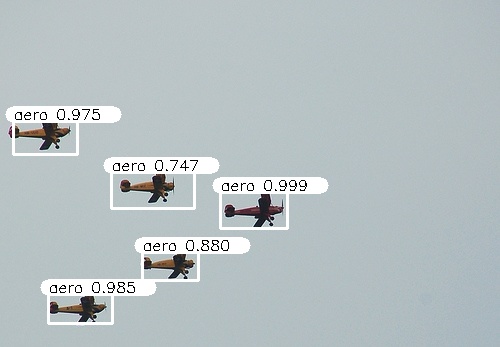}
        			\includegraphics[scale=0.15]{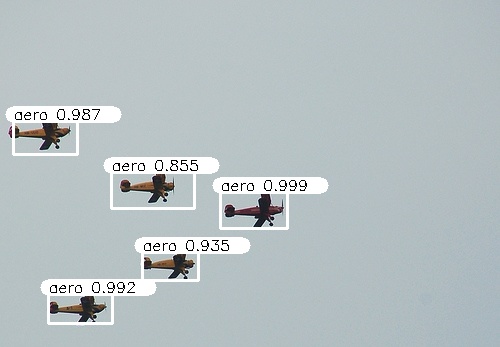}
        		
        		\caption{Visualization of hits and misses on VOC 2007 images.In each image, the first and third columns show the results of ZFnet and VGGnet. The results of our contextual model based on ZFnet and VGGnet are displayed in second and forth columns.}
        		\label{fig:show} 
        	\end{center}
        \end{figure}

Figures~\ref{fig:show} visualizes the performance of our method against Faster R-CNN on some VOC2007 images. In most situations, our method can improve the detection performance. However, the results of the image in the third line which is full of tiny boats are worse than the results of ZFnet. Actually, we find that it's because our global contextual part could not recognize the scene as a lake well. And it reminds us there still has very large development space.

\section{Conclusions}\label{section:conclusions}

In this paper, we have proposed an ensamble object detection system which combines the local apperance, the contextual relationships among different objects and the scene context information. Here, we leverage the powerful deep convolutional neural networks to obtain unary potentials for object proposals and extract features representing scene context. In addition, the pairwise potentials which take both semantic and spatial relevance into account for different object proposals are utilized to produce a semantically coherent interpretation of the input image. We fomulate the whole problem in the form of a fully-connected CRF model which can be efficiently solved by a fast mean field inference method. Furthermore, our experimental evaluation has demonstrated that our approach could effectively leverage the contextual information to improve detection accuracy, thus outperforming existing detection techniques on benchmark datasets. In the future, we will devise a better pairwise model based on CNN and incorporate it into an end-to-end framework.

\bibliographystyle{splncs}
\bibliography{egbib}
\end{document}